\documentclass{article}

\usepackage{graphicx}                    %
\usepackage{tikz}                       %
\usepackage{mathtools} 					%
\usepackage{amssymb}   					%
\usepackage{siunitx}   					%
\usepackage{amsmath}                    %
\usepackage{amsthm}                     %
\usepackage{tabularx}                   %
\usepackage{multirow}
\usepackage{booktabs}   				%
\usepackage{longtable}                  %
\usepackage{enumitem}                   %
\usepackage{caption}                    %
\usepackage{subcaption}                 %
\usepackage{epstopdf}                   %
\usepackage[%
  vlined,%
  boxed%
]{algorithm2e}                          %

\usepackage[final]{corl_2024} %

\usepackage{wrapfig}

\title{TacEx: GelSight Tactile Simulation in Isaac Sim -- Combining Soft-Body and Visuotactile Simulators}

\author{%
Duc Huy Nguyen$^{1,3}$, Tim Schneider$^{1,2}$, Guillaume Duret$^{1,2}$, \\
\textbf{Alap Kshirsagar$^{1}$, Boris Belousov$^{3}$, Jan Peters$^{1,3,4}$} \\ \\
$^{1}$TU Darmstadt \quad $^{2}$École centrale de Lyon \quad ${^3}$DFKI \quad ${^4}$Hessian.AI \\
\texttt{duc\_huy.nguyen@dfki.de}
}

\begin{document}
\maketitle

\begin{abstract}
    Training robot policies in simulation is becoming increasingly popular; nevertheless, a precise, reliable, and easy-to-use tactile simulator for contact-rich manipulation tasks is still missing.
    To close this gap, we develop TacEx -- a modular tactile simulation framework.
    We embed a state-of-the-art soft-body simulator for contacts named GIPC and vision-based tactile simulators Taxim and FOTS into Isaac Sim to achieve robust and plausible simulation of the visuotactile sensor GelSight Mini.
    We implement several Isaac Lab environments for Reinforcement Learning (RL) leveraging our TacEx simulation, including object pushing, lifting, and pole balancing.
    We validate that the simulation is stable and that the high-dimensional observations, such as the gel deformation and the RGB images from the GelSight camera, can be used for training.
    The code, videos, and additional results will be released online \url{https://sites.google.com/view/tacex}.
\end{abstract}


\section{Introduction}
Tactile sensing plays an important role for human perception of touch~\cite{dahiya2009tactile} and for advanced manipulation tasks in robotics~\cite{PreciseRoboticNeedleThreadingyu2023,CableRoutingAssemblywilson2023,VisualTactilePerceptionBasedwang2024}.
Contact properties such as contact geometry, object stiffness, and surface texture can be estimated using tactile sensors~\citep{ReviewTactileInformationli2020}. 
Furthremore, 
slip detection~\citep{MeasurementShearSlipyuan2015}, hardness estimation~\citep{ShapeindependentHardnessEstimationyuan2017a}, and grasping of soft objects~\citep{LearningGeneralizableVisionTactilehan2024} are facilitated by the sense of touch.
However, finger coordination based on tactile feedback is a complex control problem with high-dimensional observation space, therefore several Deep RL approaches have been explored~\citep{DeepReinforcementLearningchurch2020,GeneralPurposeSim2RealProtocolchen2024}. A crucial bottleneck for applying RL to tactile-rich manipulation tasks is the lack of stable and reliable contact simulation that includes soft-body interaction and tactile sensing.
Although a number of simulators have appeared recently that aim to remedy this issue~\citep{GeneralPurposeSim2RealProtocolchen2024,TaximExamplebasedSimulationsi2021,TactileGymSimtoreallin2022,DIFFTACTILEPhysicsbasedDifferentiablesi2024,xu2022efficient}, each simulator uses a different physics engine, simulates a different tactile sensor, different robot, and runs in a different robotics simulator altogether -- making comparison and interoperability challenging.

To address these issues, we develop TacEx -- a novel tactile simulation framework embedded in NVIDIA's Isaac Sim~\citep{IsaacSim} and Isaac Lab~\citep{IsaacLab,mittal2023orbit} that is modular, extensible, and based on the latest advancements in tactile simulation. We additionally integrate GIPC~\cite{GIPC} for GPU-accelerated and inversion-free simulation of soft-body contacts.
By leveraging Isaac Sim, we gain access to powerful features, such as photorealistic rendering, ROS support, and GPU-accelerated physics simulation, and by integrating TacEx into Isaac Lab -- an extensible RL framework built on top of Isaac Sim -- we enable support for teleoperation, GPU-parallelized training, and various RL libraries.

\paragraph{Related Work}
A simulation of a GelSight tactile sensor generally requires three components: physics simulation (to capture contact properties), optical simulation (to generate perceived RGB images), and marker simulation (to generate marker motion field, which reflects gel deformation).
In this paper, we chiefly focus on the physics simulation, leveraging  existing GelSight simulators for the other components: Taxim~\cite{TaximExamplebasedSimulationsi2021} for optical simulation and FOTS~\cite{FOTSFastOpticalzhao2024} for marker motion field simulation.
For physics simulation, PyBullet's rigid body dynamics has been used in TACTO~\cite{TACTOFastFlexiblewang2022} and in~\cite{TactileSimtoRealPolicychurch}, whereas~\cite{xu2022efficient} used a penalty-based contact model to 
approximate the soft gelpad deformation with rigid body dynamics.
Though fast, these methods are less accurate compared to Finite Element Methods (FEM).
More recent approaches rely on FEM: 
TacIPC~\cite{TacIPCIntersectionInversionFreedu2024} and \cite{GeneralPurposeSim2RealProtocolchen2024}
use the incremental potential contact (IPC)~\citep{IncrementalPotentialContactli2020} model to simulate the
gelpad deformation in an FEM-like manner.
DiffTactile~\cite{DIFFTACTILEPhysicsbasedDifferentiablesi2024} also simulates the gelpad deformation with an FEM-based approach.
Furthermore, the FEM simulation of Isaac Gym's Flex engine has also been used for accurate tactile simulation~\cite{TactileImprintSimulationcui2023,duret}, but it is slow and unsuited for RL.

Our approach is closest to the concurrent work TacSL~\cite{akinola2024tacsl} which also incorporates visuotactile simulation into Isaac Sim, however in contrast to TacSL, we leverage GIPC~\cite{GIPC} for FEM-based soft-body simulation (instead of a simplified soft contact model). GIPC additionally allows for soft-to-soft contact simulation.
Importantly, our method is modular, enabling the user to select which simulations should be enabled depending on the task requirements.

\section{TacEx Simulation Framework}
\label{sec:framework_details}
\begin{figure}[t]
	\centering
	\includegraphics[width=0.8\textwidth]{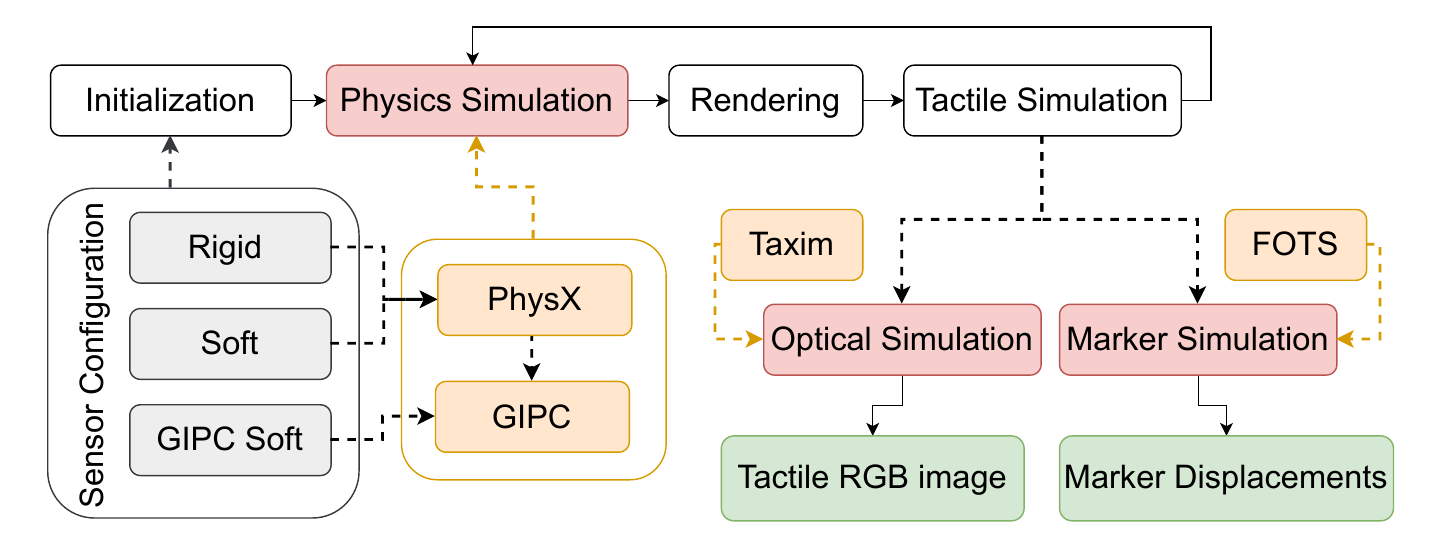}
	\caption{
		\textbf{Overview of the TacEx Tactile Simulation Pipeline}.
        First, the simulation is initialized according to a given Sensor Configuration.
		Then the physics are simulated using PhysX and GIPC, followed by the scene rendering.
        Finally, the tactile sensor is simulated using
		the optical simulation (Taxim) and marker simulation (FOTS),
        yielding a tactile RGB image and a marker displacements field.
        After this, the physics are simulated again and the process repeats.
	}
	\label{fig:tactile_sim_pipeline}
    \vspace{-1em}
\end{figure}
In this section, we present TacEx -- our modular framework for tactile simulation (c.f., Fig.~\ref{fig:tactile_sim_pipeline}).
We compare three different approaches for simulating the physical behavior of sensors and objects:
i)~PhysX to simulate the gelpad as a rigid body with compliant contact;
ii) PhysX FEM-based soft body simulation for the gelpad; 
iii) GIPC~\cite{GIPC} to simulate the gelpad as a soft body.
Since PhysX is the built-in physics engine of Isaac Sim, baselines i) and ii) are straightforward to implement by directly setting asset properties; for iii), we modified the GIPC code and created Python bindings.

We integrate the GIPC simulation with Isaac Sim in the following manner.
Isaac Sim is used for scene setup, robot simulation, and rendering.
The gelpad is attached to the sensor case and moves kinematically in response to the robot's motion, which is handled by PhysX.
The non-attached gelpad vertices are handled by GIPC: as the robot moves in Isaac Sim, the sensor case moves and the attachment points are recomputed, followed by a call to the GIPC solver that computes new positions for the remaining gelpad vertices and other GIPC-modeled objects.
This enables the gelpad and objects to move, deform, and interact dynamically in Isaac Sim.

For optical simulation, we use Taxim~\citep{TaximExamplebasedSimulationsi2021};
specifically, a GPU-accelerated implementation~\citep{TaximGPU}.
    First, we generate height maps with cameras in Isaac Sim and smooth them with pyramid Gaussian kernels.
	Then we use a polynomial lookup table to map the surface normals of the height maps to RGB values.
	As a final step, we attach shadows to the images.
    We further use the generated height maps
	to simulate the marker motion with FOTS~\citep{FOTSFastOpticalzhao2024}.
	For this, we compute the contact centers based on the height maps and extract the \(z\) rotation
	of the objects relative to the gelpads from Isaac Sim.

\section{Demonstrations and Evaluation}
\label{sec:demonstartions_and_eval}

\begin{figure}[t]
	\centering
	\includegraphics[width=\textwidth]{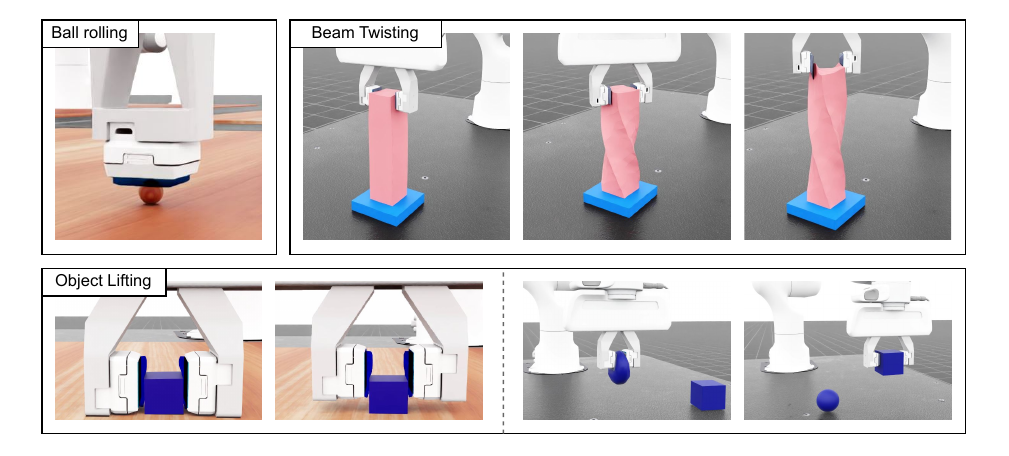}
	\caption{
		\textbf{Simulation Showcases}. We do a ball rolling experiment for testing the simulation performance, and we further
        evaluate the capabalities of our GIPC simulation by twisting and stretching a soft body beam and
        test how well the gelpads can be used for lifting objects (see website for videos).
	}
	\label{fig:showcases}
 \vspace{-1em}
\end{figure}

	We showcase the behavior, capabilities, and limitations of our framework in a series of experiments (visualized in~\autoref{fig:showcases}).
	First, a ball rolling experiment demonstrates a contact-rich manipulation task with a single GelSight sensor.
	Second, object lifting with two soft gelpads showcases robust grasping capabilities.
	Third, the limits of GIPC simulation are tested in a challenging beam twisting environment.
	Subsequently, we implement three RL tasks: object pushing, object lifting, and pole balancing -- to demonstrate how TacEx can be used within Isaac Lab for RL training.
    
\textbf{Ball Rolling.} %
	To showcase how the simulation behaves in a dynamic setting, we do a ball-rolling experiment, similar to
	\citep{xu2022efficient}.
	A robot with a single GelSight Mini sensor uses the gelpad to roll the ball around.
	For this, we define goal positions for the end-effector and compute the required joint values
	with differential inverse kinematics.
	We can simulate about $18$ robots at the same time with the rigid body configuration till we reach our VRAM limits due to camera simulations requiring much memory.
	When using soft-body GIPC-based simulation, we can only simulate a single robot properly due to our VRAM limit.
	
\textbf{Object Lifting.}
\label{para:obj_lift}
	In this example, we try to grasp and lift primitive objects using two GelSight sensors.
	The example reveals that the PhysX soft body setup cannot be used to grasp and lift objects.
	The objects always slip away even with several variations of the soft body parameters.
	Also, using a single soft body gelpad and a rigid body gelpad is unreliable for grasping and lifting.
	One reason for this failure is that the soft body simulation of PhysX currently does not support static friction.

\textbf{Beam Twisting.}
	We use this example to showcase the capabilities of the GIPC soft body simulation.
	A beam is simulated as a soft body and attached to a plate. 
	The robot grasps the top of the beam with two soft body gelpads, twists and stretches the beam till it snaps back.
	This environment demonstrates that the simulation stays stable even under extreme deformations.
	Additionally, it showcases that friction is reasonably simulated.
 
 \begin{wrapfigure}{r}{0.5\textwidth}
	\centering
    \vspace{-3em}
	\includegraphics[width=0.45\textwidth]{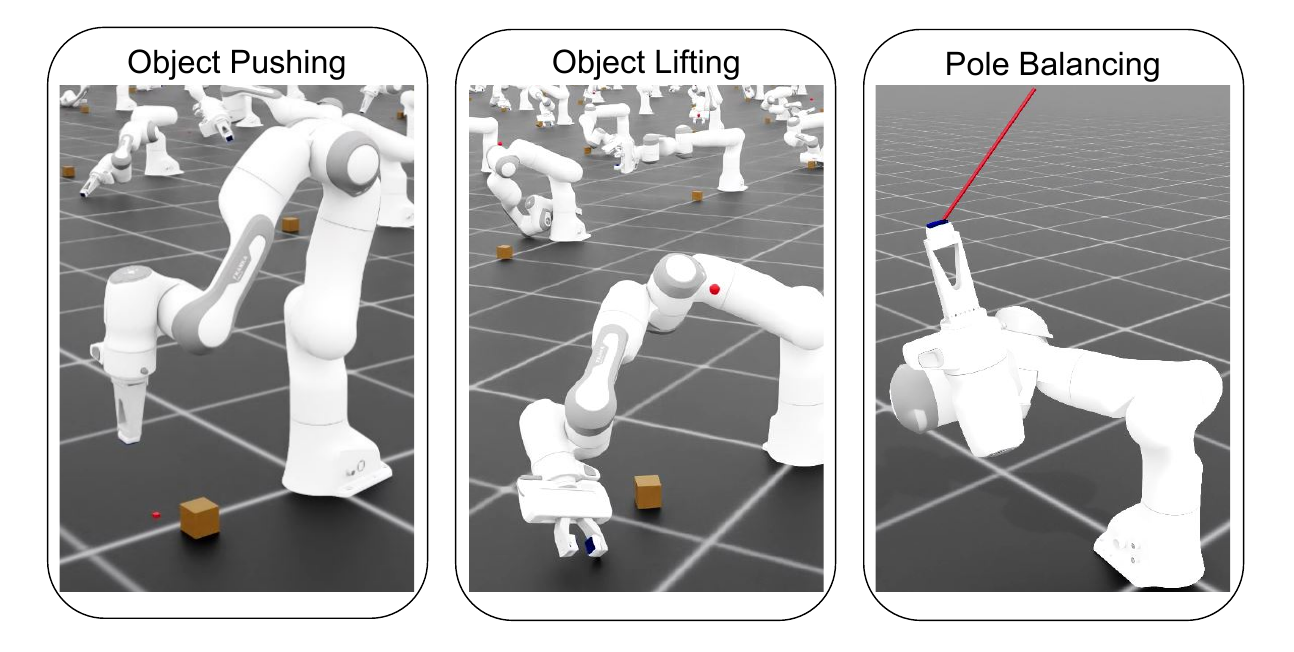}
    \vspace{-2em}
\end{wrapfigure}
\textbf{RL environments.}
	We implemented 3 environments in Isaac Lab and trained policies to validate that our framework can be used for Reinforcement Learning.
    In each environment, we used the marker displacements from our tactile simulation.
    For training the RL policies, we used PPO~\cite{schulman2017proximalpolicyoptimizationalgorithms} as implemented in~\cite{schneider2023learning}.
    We have validated that the training pipeline works, and we are currently working towards obtaining successful policies that leverage tactile feedback.

\begin{table}[t]
    \caption{\textbf{Tactile Simulation Speed.} We measure the average simulation time per frame in $\mathrm{ms}$ for optical (GPU accelerated Taxim with shadows) and for marker simulation (FOTS running on CPU without parallelization) during contact in the ball rolling experiment with rigid body gelpads.
    We additionally measure the performance of the height map generation with the Isaac Sim USD cameras.
    We generate tactile RGB images with a resolution of $480 \times 640$ and simulate $10 \times 10$ markers per environment. %
    We were only able to test up to approximately $18$ environments, else we run into out-of-memory issues. We assume that the performance loss at $16$ environments is related to our GPU being near its limit.}
    \centering
    \vspace{0.5em}
    \begin{tabular}{c||c|c|c} 
     num\_envs & height map gen & optical sim & marker sim \\ 
     \hline
     1 & 1.3718 & 5.9015 & 4.4863 \\ 
     2 & 0.8508 & 3.8886 & 2.8838  \\
     4 & 0.5988 & 3.0424 & 2.1184 \\
     8 & 0.4323 & 2.5773 & 1.7587 \\
     16 & 2.8827 & 5.7314 & 5.0450 \\ 
     18 & 3.5149 & 5.931 & 5.2343 \\ 
    \end{tabular}
    \label{table:1}
\end{table}
\begin{table}[t]
    \caption{\textbf{PhysX Simulation Speed.} We measure the average simulation time per frame in $\mathrm{ms}$ for the physics simulation with PhysX during the ball rolling experiment (without tactile simulation). We run into out-of-memory issues with the soft body simulation at 256 environments. The soft body gelpad has a mesh resolution of $10$ and uses $16$ solver iterations.} %
    \centering
    \vspace{0.5em}
    \begin{tabular}{c||c|c|c|c|c|c|c|c} 
     num\_envs & 1 & 16 & 32 & 64 & 128 & 256 & 512 & 1024 \\ 
     \hline
     rigid & 3.6930 & 0.2426 & 0.1286 & 0.0673 & 0.0361 & 0.0212 & 0.0143 & 0.0093\\ 
     \hline
     soft &  4.7069 & 0.4496 & 0.2718 & 0.1798 & 0.1267 & - & - & - \\
    \end{tabular}
    \label{table:2}
\end{table}
\begin{wraptable}{r}{0.4\textwidth}
    \caption{\textbf{GIPC Simulation Speed.}}
    \centering
    \begin{tabular}{c c|c} 
     num\_vert & num\_tetra &  GIPC \\ 
     \hline
     1029 & 3717 & 24.95 $\mathrm{ms}$ \\ 
     7900 & 40370 & 110.47  $\mathrm{ms}$\\
     12509 & 66563 & 221.61 $\mathrm{ms}$ \\
    \end{tabular}
    \vspace{-1em}
    \label{table:3}
\end{wraptable}
\textbf{Speed Evaluations.} We evaluate the simulation time of the optical simulation Taxim/FOTS in Table~\ref{table:1}.
Results for PhysX are presented in Table~\ref{table:2}.
The runtimes for soft-body GIPC simulation are shown in Table~\ref{table:3}.
We measure the average simulation time per frame in $\mathrm{ms}$ for the physics simulation with GIPC during the ball rolling experiment without tactile simulation. Compared to the ball rolling experiments with PhysX, the ball here is a soft body.
We use different mesh resolutions for the ball to measure the performance w.r.t the amount of vertices and tetrahedra.

\vspace{-0.5em}
\section{Conclusion and Future Work}
\label{sec:conclusion}
	We presented TacEx -- a novel framework for simulating GelSight tactile sensors.
	The framework enables the usage of GelSight Mini sensors for Reinforcement Learning.
	It is built on top of Isaac Sim and Isaac Lab, which gives the user access to a wealth of features non-existent in 
	current tactile simulators.
	We designed the framework to be modular, extendable, and easy to use.
	The framework integrates multiple different simulation approaches.
	The gelpad can either be simulated as a rigid body with compliant contact, or as a soft body.
	For the soft body simulation, one can use PhysX or our integration of GIPC.
	To simulate the sensor output, we create height maps with cameras in Isaac Sim and use the approach from
	Taxim~\cite{TaximExamplebasedSimulationsi2021} for the optical and the one from FOTS~\cite{FOTSFastOpticalzhao2024} for the marker simulation.
	We demonstrated framework features and simulation behavior with multiple examples.
	Additionally, we provide three environments for RL with tactile sensing.

	A limitation of our current work is that it only contains qualitative experiments and demonstrations in simulation.
    Our tactile simulation, specifically the physics simulation approaches, lacks experiments that investigate whether they can be used for Sim2Real or not.
	Therefore, we aim to do more quantitative experiments for comparing different tactile simulation approaches,
	as well as Sim2Real experiments in future works.
	We also plan to extend our framework to include more RL environments and more tactile simulation approaches with the goal of creating a benchmarking platform for tactile simulators and algorithms for tactile sensing.

\acknowledgments{This research was supported by Research Clusters “The Adaptive Mind” and “Third Wave of AI”, funded by the Excellence Program of the Hessian Ministry of Higher Education, Science, Research and the Arts. This work was supported by the French Research Agency, l’Agence Nationale de Recherche (ANR), and the German Federal Ministry of Education and Research (BMBF) through the project Aristotle (ANR-21-FAI1-0009-01).
The authors acknowledge the grant ``Einrichtung eines Labors des Deutschen Forschungszentrum für Künstliche Intelligenz (DFKI) an der Technischen Universität Darmstadt'' of the Hessisches Ministerium für Wissenschaft und Kunst.
The authors gratefully acknowledge the computing time provided to them on the high-performance computer Lichtenberg II at TU Darmstadt, funded by the BMBF and the State of Hesse.}

\bibliography{example.bib}  %

\begin{thebibliography}{36}
\providecommand{\natexlab}[1]{#1}
\providecommand{\url}[1]{\texttt{#1}}
\expandafter\ifx\csname urlstyle\endcsname\relax
  \providecommand{\doi}[1]{doi: #1}\else
  \providecommand{\doi}{doi: \begingroup \urlstyle{rm}\Url}\fi

\bibitem[Dahiya et~al.(2009)Dahiya, Metta, Valle, and Sandini]{dahiya2009tactile}
R.~S. Dahiya, G.~Metta, M.~Valle, and G.~Sandini.
\newblock Tactile sensing—from humans to humanoids.
\newblock \emph{IEEE transactions on robotics}, 26\penalty0 (1):\penalty0 1--20, 2009.

\bibitem[Yu et~al.()Yu, Xu, Yao, Ren, Tang, Li, Gu, and Lu]{PreciseRoboticNeedleThreadingyu2023}
Z.~Yu, W.~Xu, S.~Yao, J.~Ren, T.~Tang, Y.~Li, G.~Gu, and C.~Lu.
\newblock Precise {{Robotic Needle-Threading}} with {{Tactile Perception}} and {{Reinforcement Learning}}.

\bibitem[Wilson et~al.()Wilson, Jiang, Lian, and Yuan]{CableRoutingAssemblywilson2023}
A.~Wilson, H.~Jiang, W.~Lian, and W.~Yuan.
\newblock Cable {{Routing}} and {{Assembly}} using {{Tactile-driven Motion Primitives}}.

\bibitem[Wang et~al.()Wang, Liu, Liu, Huang, and Yang]{VisualTactilePerceptionBasedwang2024}
G.~Wang, X.~Liu, Z.~Liu, P.~Huang, and Y.~Yang.
\newblock Visual-{{Tactile Perception Based Control Strategy}} for {{Complex Robot Peg-in-Hole Process}} via {{Topological}} and {{Geometric Reasoning}}.
\newblock 9\penalty0 (10):\penalty0 8410--8417.
\newblock ISSN 2377-3766.
\newblock \doi{10.1109/LRA.2024.3436334}.

\bibitem[Li et~al.()Li, Kroemer, Su, Veiga, Kaboli, and Ritter]{ReviewTactileInformationli2020}
Q.~Li, O.~Kroemer, Z.~Su, F.~F. Veiga, M.~Kaboli, and H.~J. Ritter.
\newblock A review of tactile information: Perception and action through touch.
\newblock 36\penalty0 (6):\penalty0 1619--1634.
\newblock ISSN 1941-0468.
\newblock \doi{10.1109/TRO.2020.3003230}.
\newblock URL \url{https://ieeexplore.ieee.org/document/9136877/?arnumber=9136877}.
\newblock Conference Name: {IEEE} Transactions on Robotics.

\bibitem[Yuan et~al.({\natexlab{a}})Yuan, Li, Srinivasan, and Adelson]{MeasurementShearSlipyuan2015}
W.~Yuan, R.~Li, M.~A. Srinivasan, and E.~H. Adelson.
\newblock Measurement of shear and slip with a {{GelSight}} tactile sensor.
\newblock In \emph{2015 {{IEEE International Conference}} on {{Robotics}} and {{Automation}} ({{ICRA}})}, pages 304--311. IEEE, {\natexlab{a}}.
\newblock ISBN 978-1-4799-6923-4.
\newblock \doi{10.1109/ICRA.2015.7139016}.

\bibitem[Yuan et~al.({\natexlab{b}})Yuan, Zhu, Owens, Srinivasan, and Adelson]{ShapeindependentHardnessEstimationyuan2017a}
W.~Yuan, C.~Zhu, A.~Owens, M.~A. Srinivasan, and E.~H. Adelson.
\newblock Shape-independent hardness estimation using deep learning and a {GelSight} tactile sensor.
\newblock In \emph{2017 {IEEE} International Conference on Robotics and Automation ({ICRA})}, pages 951--958, {\natexlab{b}}.
\newblock \doi{10.1109/ICRA.2017.7989116}.
\newblock URL \url{http://arxiv.org/abs/1704.03955}.

\bibitem[Han et~al.()Han, Yu, Batra, Boyd, Mehta, Zhao, She, Hutchinson, and Zhao]{LearningGeneralizableVisionTactilehan2024}
Y.~Han, K.~Yu, R.~Batra, N.~Boyd, C.~Mehta, T.~Zhao, Y.~She, S.~Hutchinson, and Y.~Zhao.
\newblock Learning generalizable vision-tactile robotic grasping strategy for deformable objects via transformer.
\newblock pages 1--13.
\newblock ISSN 1941-014X.
\newblock \doi{10.1109/TMECH.2024.3400789}.
\newblock URL \url{https://ieeexplore.ieee.org/document/10552075/?arnumber=10552075}.
\newblock Conference Name: {IEEE}/{ASME} Transactions on Mechatronics.

\bibitem[Church et~al.()Church, Lloyd, Hadsell, and Lepora]{DeepReinforcementLearningchurch2020}
A.~Church, J.~Lloyd, R.~Hadsell, and N.~F. Lepora.
\newblock Deep {{Reinforcement Learning}} for {{Tactile Robotics}}: {{Learning}} to {{Type}} on a {{Braille Keyboard}}.
\newblock 5\penalty0 (4):\penalty0 6145--6152.
\newblock ISSN 2377-3766.
\newblock \doi{10.1109/LRA.2020.3010461}.

\bibitem[Chen et~al.()Chen, Xu, Xiang, Yuan, Su, and Chen]{GeneralPurposeSim2RealProtocolchen2024}
W.~Chen, J.~Xu, F.~Xiang, X.~Yuan, H.~Su, and R.~Chen.
\newblock General-{{Purpose Sim2Real Protocol}} for {{Learning Contact-Rich Manipulation With Marker-Based Visuotactile Sensors}}.
\newblock 40:\penalty0 1509--1526.
\newblock ISSN 1552-3098, 1941-0468.
\newblock \doi{10.1109/TRO.2024.3352969}.

\bibitem[Si and Yuan()]{TaximExamplebasedSimulationsi2021}
Z.~Si and W.~Yuan.
\newblock Taxim: {{An Example-based Simulation Model}} for {{GelSight Tactile Sensors}}.

\bibitem[Lin et~al.()Lin, Lloyd, Church, and Lepora]{TactileGymSimtoreallin2022}
Y.~Lin, J.~Lloyd, A.~Church, and N.~F. Lepora.
\newblock Tactile {{Gym}} 2.0: {{Sim-to-real Deep Reinforcement Learning}} for {{Comparing Low-cost High-Resolution Robot Touch}}.

\bibitem[Si et~al.()Si, Zhang, Ben, Romero, Xian, Liu, and Gan]{DIFFTACTILEPhysicsbasedDifferentiablesi2024}
Z.~Si, G.~Zhang, Q.~Ben, B.~Romero, Z.~Xian, C.~Liu, and C.~Gan.
\newblock {{DIFFTACTILE}}: {{A Physics-based Differentiable Tactile Simulator}} for {{Contact-rich Robotic Manipulation}}.

\bibitem[Xu et~al.()Xu, Kim, Chen, Garcia, Agrawal, Matusik, and Sueda]{xu2022efficient}
J.~Xu, S.~Kim, T.~Chen, A.~R. Garcia, P.~Agrawal, W.~Matusik, and S.~Sueda.
\newblock Efficient tactile simulation with differentiability for robotic manipulation.
\newblock In \emph{6th Annual Conference on Robot Learning}.

\bibitem[NVIDIA({\natexlab{a}})]{IsaacSim}
NVIDIA.
\newblock Nvidia isaac sim, {\natexlab{a}}.
\newblock URL \url{https://developer.nvidia.com/isaac/sim}.

\bibitem[NVIDIA({\natexlab{b}})]{IsaacLab}
NVIDIA.
\newblock Isaac lab, {\natexlab{b}}.
\newblock URL \url{https://isaac-sim.github.io/IsaacLab/}.

\bibitem[Mittal et~al.(2023)Mittal, Yu, Yu, Liu, Rudin, Hoeller, Yuan, Singh, Guo, Mazhar, Mandlekar, Babich, State, Hutter, and Garg]{mittal2023orbit}
M.~Mittal, C.~Yu, Q.~Yu, J.~Liu, N.~Rudin, D.~Hoeller, J.~L. Yuan, R.~Singh, Y.~Guo, H.~Mazhar, A.~Mandlekar, B.~Babich, G.~State, M.~Hutter, and A.~Garg.
\newblock Orbit: A unified simulation framework for interactive robot learning environments.
\newblock \emph{IEEE Robotics and Automation Letters}, 8\penalty0 (6):\penalty0 3740--3747, 2023.
\newblock \doi{10.1109/LRA.2023.3270034}.

\bibitem[Huang et~al.(2024)Huang, Chitalu, Lin, and Komura]{GIPC}
K.~Huang, F.~M. Chitalu, H.~Lin, and T.~Komura.
\newblock Gipc: Fast and stable gauss-newton optimization of ipc barrier energy.
\newblock \emph{ACM Trans. Graph.}, 43\penalty0 (2), 3 2024.
\newblock ISSN 0730-0301.
\newblock \doi{10.1145/3643028}.
\newblock URL \url{https://doi.org/10.1145/3643028}.

\bibitem[Zhao et~al.()Zhao, Qian, Duan, and Luo]{FOTSFastOpticalzhao2024}
Y.~Zhao, K.~Qian, B.~Duan, and S.~Luo.
\newblock {{FOTS}}: {{A Fast Optical Tactile Simulator}} for {{Sim2Real Learning}} of {{Tactile-motor Robot Manipulation Skills}}.

\bibitem[Wang et~al.()Wang, Lambeta, Chou, and Calandra]{TACTOFastFlexiblewang2022}
S.~Wang, M.~Lambeta, P.-W. Chou, and R.~Calandra.
\newblock {{TACTO}}: {{A Fast}}, {{Flexible}}, and {{Open-source Simulator}} for {{High-Resolution Vision-based Tactile Sensors}}.
\newblock 7\penalty0 (2):\penalty0 3930--3937.
\newblock ISSN 2377-3766, 2377-3774.
\newblock \doi{10.1109/LRA.2022.3146945}.

\bibitem[Church and Lloyd()]{TactileSimtoRealPolicychurch}
A.~Church and J.~Lloyd.
\newblock Tactile {{Sim-to-Real Policy Transfer}} via {{Real-to-Sim Image Translation}}.

\bibitem[Du et~al.()Du, Xu, Ren, Yu, and Lu]{TacIPCIntersectionInversionFreedu2024}
W.~Du, W.~Xu, J.~Ren, Z.~Yu, and C.~Lu.
\newblock {{TacIPC}}: {{Intersection-}} and {{Inversion-Free FEM-Based Elastomer Simulation}} for {{Optical Tactile Sensors}}.
\newblock 9\penalty0 (3):\penalty0 2559--2566.
\newblock ISSN 2377-3766, 2377-3774.
\newblock \doi{10.1109/LRA.2024.3357030}.

\bibitem[Li et~al.()Li, Ferguson, Schneider, Langlois, Zorin, Panozzo, Jiang, and Kaufman]{IncrementalPotentialContactli2020}
M.~Li, Z.~Ferguson, T.~Schneider, T.~Langlois, D.~Zorin, D.~Panozzo, C.~Jiang, and D.~M. Kaufman.
\newblock Incremental potential contact: Intersection-and inversion-free, large-deformation dynamics.
\newblock 39\penalty0 (4).
\newblock ISSN 0730-0301, 1557-7368.
\newblock \doi{10.1145/3386569.3392425}.

\bibitem[Cui et~al.()Cui, Wang, Wang, Li, Wang, and Zhang]{TactileImprintSimulationcui2023}
S.~Cui, Y.~Wang, S.~Wang, Q.~Li, R.~Wang, and C.~Zhang.
\newblock Tactile {{Imprint Simulation}} of {{GelStereo Visuotactile Sensors}}.
\newblock In \emph{2023 {{IEEE International Conference}} on {{Mechatronics}} and {{Automation}} ({{ICMA}})}, pages 650--656. IEEE.
\newblock ISBN 9798350320848.
\newblock \doi{10.1109/ICMA57826.2023.10216245}.

\bibitem[Duret et~al.(2024)Duret, Zara, Peters, and Chen]{duret}
G.~Duret, F.~Zara, J.~Peters, and L.~Chen.
\newblock {Toward synthetic data generation for robotic tactile manipulations}.
\newblock In \emph{{Workshop on ''Robot Embodiment through Visuo-Tactile Perception'' - 2024 IEEE International Conference on Robotics and Automation (ICRA) Conference Workshop}}, Yokohama, Japan, May 2024.
\newblock URL \url{https://hal.science/hal-04566202}.

\bibitem[Akinola et~al.(2024)Akinola, Xu, Carius, Fox, and Narang]{akinola2024tacsl}
I.~Akinola, J.~Xu, J.~Carius, D.~Fox, and Y.~Narang.
\newblock Tacsl: A library for visuotactile sensor simulation and learning.
\newblock \emph{arXiv preprint arXiv:2408.06506}, 2024.

\bibitem[Schneider()]{TaximGPU}
T.~Schneider.
\newblock Taxim-gpu.
\newblock URL \url{https://git.ias.informatik.tu-darmstadt.de/tactile-sensing/taxim-gpu}.

\bibitem[Schulman et~al.()Schulman, Wolski, Dhariwal, Radford, and Klimov]{schulman2017proximalpolicyoptimizationalgorithms}
J.~Schulman, F.~Wolski, P.~Dhariwal, A.~Radford, and O.~Klimov.
\newblock Proximal policy optimization algorithms.
\newblock URL \url{https://arxiv.org/abs/1707.06347}.

\bibitem[Schneider et~al.(2023)Schneider, Frey, Miki, and Hutter]{schneider2023learning}
L.~Schneider, J.~Frey, T.~Miki, and M.~Hutter.
\newblock Learning risk-aware quadrupedal locomotion using distributional reinforcement learning, 2023.

\bibitem[Higuera et~al.()Higuera, Boots, and Mukadam]{LearningReadBraillehiguera2023}
C.~Higuera, B.~Boots, and M.~Mukadam.
\newblock Learning to {{Read Braille}}: {{Bridging}} the {{Tactile Reality Gap}} with {{Diffusion Models}}.

\bibitem[Zhong et~al.()Zhong, Albini, Jones, Maiolino, and Posner]{TouchingNeRFLeveragingzhong}
S.~Zhong, A.~Albini, O.~P. Jones, P.~Maiolino, and I.~Posner.
\newblock Touching a {{NeRF}}: {{Leveraging Neural Radiance Fields}} for {{Tactile Sensory Data Generation}}.

\bibitem[Kim et~al.()Kim, Yang, Kim, Kim, Kim, and Kim]{MarkerEmbeddedTactileImagekim2023}
W.~D. Kim, S.~Yang, W.~Kim, J.-J. Kim, C.-H. Kim, and J.~Kim.
\newblock Marker-{{Embedded Tactile Image Generation}} via {{Generative Adversarial Networks}}.
\newblock 8\penalty0 (8):\penalty0 4481--4488.
\newblock ISSN 2377-3766, 2377-3774.
\newblock \doi{10.1109/LRA.2023.3284370}.

\bibitem[Mehta et~al.()Mehta, Diaz, Golemo, Pal, and Paull]{ActiveDomainRandomizationmehta}
B.~Mehta, M.~Diaz, F.~Golemo, C.~J. Pal, and L.~Paull.
\newblock Active {{Domain Randomization}}.

\bibitem[Tobin et~al.()Tobin, Fong, Ray, Schneider, Zaremba, and Abbeel]{DomainRandomizationTransferringtobin2017}
J.~Tobin, R.~Fong, A.~Ray, J.~Schneider, W.~Zaremba, and P.~Abbeel.
\newblock Domain randomization for transferring deep neural networks from simulation to the real world.
\newblock In \emph{2017 {{IEEE}}/{{RSJ International Conference}} on {{Intelligent Robots}} and {{Systems}} ({{IROS}})}, pages 23--30.
\newblock \doi{10.1109/IROS.2017.8202133}.

\bibitem[Hu et~al.(2020)Hu, Schneider, Wang, Zorin, and Panozzo]{wildmeshing}
Y.~Hu, T.~Schneider, B.~Wang, D.~Zorin, and D.~Panozzo.
\newblock Fast tetrahedral meshing in the wild.
\newblock \emph{ACM Trans. Graph.}, 39\penalty0 (4), July 2020.
\newblock ISSN 0730-0301.
\newblock \doi{10.1145/3386569.3392385}.
\newblock URL \url{https://doi.org/10.1145/3386569.3392385}.

\bibitem[NVIDIA()]{USDRT}
NVIDIA.
\newblock Usd, fabric and usdrt.
\newblock URL \url{https://docs.omniverse.nvidia.com/kit/docs/usdrt/latest/docs/usd_fabric_usdrt.html}.

\end{thebibliography}

\clearpage
\appendix

\section{Related Work}

A simulation of a GelSight tactile sensor generally requires three components: physics simulation (to capture contact properties), optical simulation (to generate perceived RGB images), and marker simulation (to generate marker motion field, which reflects gel deformation).
This section provides a general overview of the approaches used by tactile simulators.

\paragraph{Physics Simulation}
Rigid body based approaches allow for fast simulations. 
TACTO~\citep{TACTOFastFlexiblewang2022} and \citet{TactileSimtoRealPolicychurch} use PyBullet's rigid body simulation. %
\citet{xu2022efficient} use a penalty-based contact model to 
approximate the soft gelpad's deformation with rigid body dynamics.
While fast, they are inaccurate compared to Finite Element Methods (FEM).
DiffTactile~\citep{DIFFTACTILEPhysicsbasedDifferentiablesi2024} simulates the gelpad deformation with a FEM-based approach.
\citet{GeneralPurposeSim2RealProtocolchen2024} and TacIPC~\citep{TacIPCIntersectionInversionFreedu2024}
use incremental potential contact (IPC)~\citep{IncrementalPotentialContactli2020} to simulate the
gelpad deformation in a FEM-based manner. %
Additionally, the FEM simulation of Isaac Gym's Flex engine also has been used for tactile simulation~\citep{TactileImprintSimulationcui2023,duret}.

\paragraph{Optical Simulation}
Common for optical simulation approaches is the generation of a height map, which represents the surface of the gelpad and its deformation after contact.
The gelpad deformation is approximated by smoothing the height map with, for example, pyramid Gaussian kernels, and then surface normals of the height map are mapped to RGB values (Taxim~\citep{TaximExamplebasedSimulationsi2021}, DiffTactile~\citep{DIFFTACTILEPhysicsbasedDifferentiablesi2024}, and FOTS~\citep{FOTSFastOpticalzhao2024}).
Taxim~uses a polynomial look-up table to map surface normals to RGB values. 
DiffTactile and FOTS use trained MLPs for the mapping.
These methods need relatively few real tactile images to work, compared to \citet{LearningReadBraillehiguera2023}, for example.
Here, tactile RGB images are generated via a diffusion model.
Another example of a data-intensive approach is the work from \citet{TouchingNeRFLeveragingzhong}, which uses Neural Radiance Fields and a conditional Generative Adversarial Network.

\paragraph{Marker Simulation}
\citet{GeneralPurposeSim2RealProtocolchen2024} and DiffTactile~\citep{DIFFTACTILEPhysicsbasedDifferentiablesi2024} use their accurate soft body simulation of the gelpad 
for the marker motion simulation. 
A mapping, which relates markers to faces of the tetrahedra mesh, is precomputed.
If the gelpad deforms, the new marker world coordinates can be computed according to the vertex positions of the corresponding facet.
A marker image is then created by projecting the marker world positions onto the image plane of a camera.
\citet{xu2022efficient} simulate the normal and shear tactile force fields.
They compute the force fields with a penalty-based contact model.
Simulated tactile force fields and the marker motion fields of real sensors are
normalized before they are used as input for an RL policy.
FOTS~\citep{FOTSFastOpticalzhao2024} simulates the marker motion field with exponential functions, which model the marker displacement distributions for normal, shear, and twist loads. 
\citet{MarkerEmbeddedTactileImagekim2023} use a generative adversarial network that
takes a sequence of depth images as input and outputs a tactile RGB image with markers.

\section{Choice of the Simulation Methods}
\label{subsec:reasoning_methods}
For the physics simulation, we leverage PhysX.
Not only because it is the built-in physics engine of Isaac Sim, but also due its fast GPU-accelerated simulation.
We simulate the gelpad as a rigid body with compliant contacts for extremely fast tactile 
simulation. This can be extremely useful for prototyping new RL environments and algorithms, for example.
Besides that, we simulate the gelpad as a soft body to have an approach with accurate gelpad simulation.
Additionally, Isaac Sims's soft body simulation has not been used for tactile simulation yet.
In this way, our work also provides a first study of the capabilities of Isaac Sims soft body simulation
for simulating GelSight sensors.
Unfortunately, some of our initial experiments revealed that the built-in soft body simulation is currently lacking in some aspects.
We tried to grasp and pick up objects with soft body gelpads, but
the objects were constantly slipping away.
Even after extensive experimentation with different soft body parameters, this behavior did not change.
One reason is the lack of static friction in their soft body simulation.

Therefore, we also wanted to integrate an external physics simulator into our framework.
It would additionally serve as an example of the extensibility of our framework. 
But the question here is, which soft body simulation should we use?
IPC~\cite{IncrementalPotentialContactli2020} seemed to be a promising candidate.
IPC is extremely robust. It guarantees intersection and inversion free simulation of soft bodies, regardless of material parameters and severity of deformation.
This allows us to freely change parameters without worrying about unstable and inaccurate simulations. 
This is crucial for RL since one technique for closing the Sim2Real gap is domain randomization~\citep{ActiveDomainRandomizationmehta,DomainRandomizationTransferringtobin2017}.
Another benefit is that the need to fine-tune simulation parameters till reasonable behavior is achieved is omitted,
or at least significantly reduced.
IPC also simulates static and dynamic friction.
Furthermore, it has already been shown that IPC can be used for accurate tactile simulation~\citep{GeneralPurposeSim2RealProtocolchen2024}.
Instead of IPC, we use GIPC~\cite{GIPC}, a completely GPU-based variant of IPC with massive speedups.

For the optical simulation, we use the approach from Taxim~\citep{TaximExamplebasedSimulationsi2021}
and for the marker simulation FOTS~\citep{FOTSFastOpticalzhao2024}.
Both approaches are based on generating a height map, which approximates the gelpad deformation.
Generating a height map is a common step for the optical simulation.
By already having this step implemented, it is easier to integrate other optical simulation approaches.
Additionally, both approaches do not rely on accurate simulation of the gelpad.
Not only is this beneficial performance-wise but also for using different types of physics simulation.
Compared to, for instance, the simulator from \citet{GeneralPurposeSim2RealProtocolchen2024},
we can simulate the marker motion field, even with a rigid body gelpad.
Another benefit of Taxim and FOTS is that they can be easily adjusted to simulate other GelSight sensor models.
Both use a relatively simple calibration process. %

\section{Simulation Details}
In this section, we describe our GIPC integration and our tactile simulation in more detail.
Subsection~\ref{subsec:general_gipc} describes the GIPC simulation pipeline in general and 
subsection~\ref{subsec:gipc_attachments} the attachment creation.
Subsection~\ref{subsec:tactile_sim} explains the steps of the optical and marker simulation.

\subsection{General GIPC Simulation Pipeline}
\label{subsec:general_gipc}
\begin{figure}
    \centering
    \includegraphics[width=\textwidth]{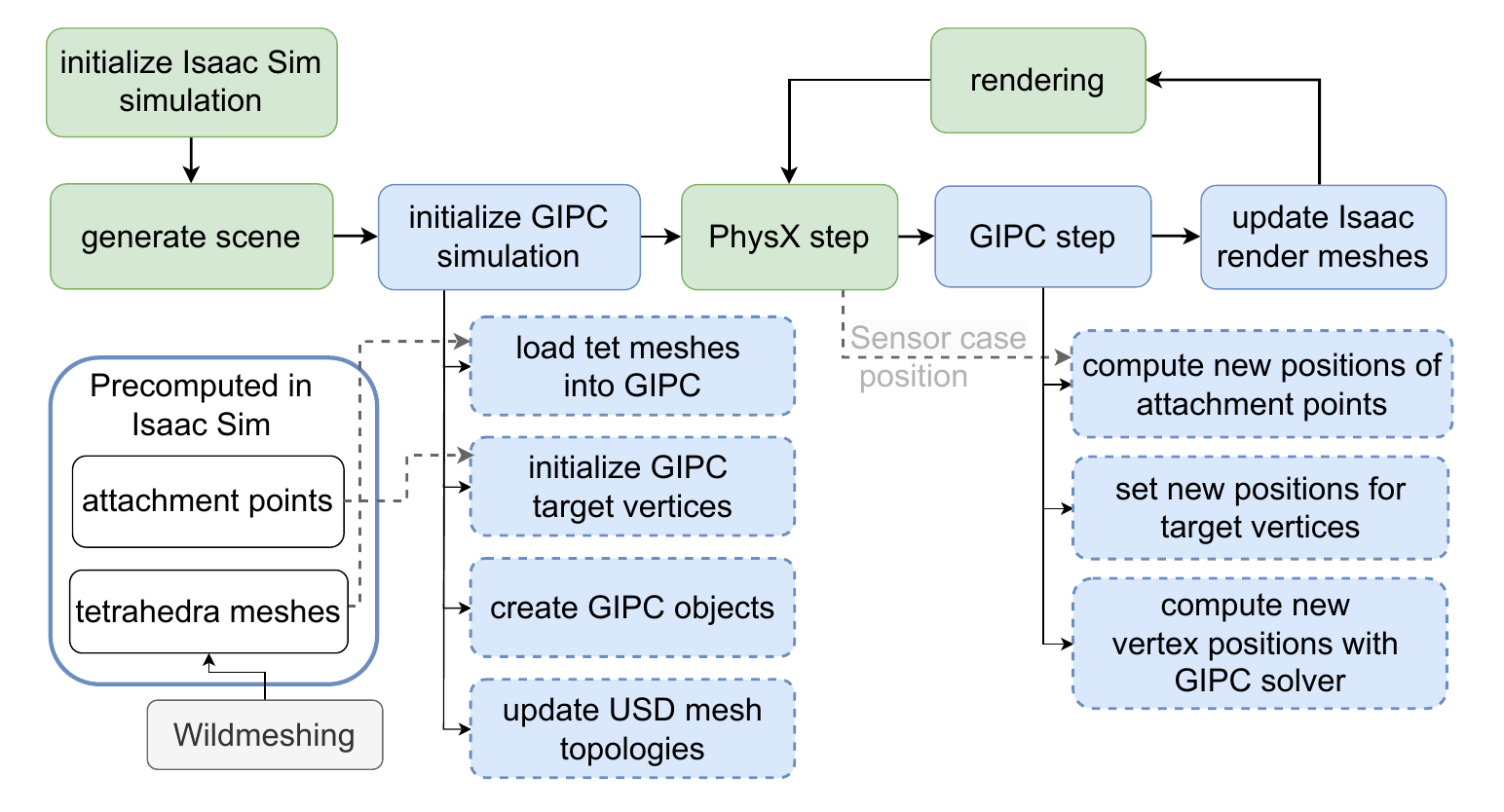}
    \caption{
    \textbf{Overview of our GIPC simulation pipeline.}
    First, the Isaac Sim simulation is initialized 
    and the scene is created. Then the GIPC simulation is initialized. This includes, loading tetrahedra meshes
    into GIPC, creating GIPC objects, and updating the corresponding meshes in Isaac Sim.
    The tetrahedra (tet) meshes are computed with Wildmeshing. Attachment points for the gelpads are also precomputed.
    For the physics simulation the simulation state after a time step is computed.
    First, with PhysX, which leads to, for example, the robot moving, and then with GIPC.
    The GIPC simulation takes the newest position of the sensor case and uses it to compute the new positions of the attachment points.
    Then the GIPC solver computes the new vertex positions of every GIPC object.
    After that, the meshes in Isaac Sim are updated correspondingly and the scene is rendered.
    }
    \label{fig:gipc_pipeline}
\end{figure}
The general simulation pipeline with GIPC looks as follows.
First, the GIPC simulation needs to be set up and initialized.
This happens after initialization of the Isaac Sim simulation and generation of the scene.
The scene generation involves spawning assets into Isaac Sim.
For our GIPC simulation, we spawn assets without physical properties into Isaac Sim and then use them to create GIPC objects.
To create a GIPC object out of an asset, we first extract the triangle mesh data of the corresponding USD mesh, i.e.,
world position and triangle indices of the mesh points.
We then generate a tetrahedra mesh, which is required for the GIPC simulation.
We use the Wildmeshing~\cite{wildmeshing} python bindings for the tetrahedra generation.
The topologies of the Isaac Sim USD meshes are updated according to the surface vertices and triangles of 
the tetrahedra meshes. This is necessary for the rendering of the objects.

After the initialization of the simulation, we compute the simulation state after a time step.
For this, we first do a PhysX step, i.e., compute the new simulation state for objects simulated by PhysX.
For instance, this involves the simulation of the robot's movement.
The PhysX step is directly followed by a GIPC step.
A step in our integrated GIPC simulation consists of first computing the new positions of attachment points.
These values are set as target vertex values for the vertices that are attachment points.
This allows us to move the GIPC objects kinematically.
At the end of the GIPC step, we compute the new vertex positions for all GIPC objects with the GIPC solver.
Additionally, we update the object position data by computing the mean of the new vertex positions.
The object position data is helpful for RL environments as observations, for example.

To render the results of the GIPC simulations inside Isaac Sim, 
we update the position of the USD mesh vertices with the new computed vertex positions.
For fast updates of the USD meshes, we use the USDRT~\citep{USDRT} API.
The rest is done by Isaac Sim's rendering engine.
The general simulation pipeline with GIPC and PhysX is visualized in~\autoref{fig:gipc_pipeline}.

For RL, we also need to reset the GIPC simulation occasionally.
Resetting means bringing the scene back to the initial state. 
To achieve this, we save the initial positions of the GIPC vertices during the scene initialization.
When the reset happens, we set the initial positions as the position of the vertices.
The velocities are set to $ (0,0,0) $.

\subsection{GIPC Attachments}
\label{subsec:gipc_attachments}
The two core questions regarding the attachment points are:
\begin{enumerate}
    \item How do we find attachment points?
    \item How do we compute the new vertex positions for the attachment points after each robot movement?
\end{enumerate}
Finding attachment points means finding the IDs of vertices that should be attachment points.
Attachment points should be the vertices inside the sensor case or at least close to it.
To attach a GIPC object to the sensor case, we first query the world position and orientation of the sensor case. 
Secondly, we iterate through each point of the GIPC object's tetrahedra mesh (tet point) and do sphere ray casting in Isaac Sim with the PhysX scene query interface.
A sphere with a specified radius is swept out from an origin point in a direction with the specified maximum.
If the sphere hits a collider mesh, the impact point is returned.
By using the world position of the tet points as the origin point, a very small sphere radius, and a very small maximum distance, we can check if 
a tet point is inside or close enough to a rigid body.
This way, we find the attachment points. %

For computing the new positions of the attachment points, we use that the relative positions of the attachment points
to the sensor case position are constant.
We precompute and save the offsets between attachment points and sensor case position.
These offsets stay the same throughout the simulation.
The attachment points positions are then updated by first querying the current pose, i.e., position and orientation, of the rigid body.
Then the attachment offsets are transformed based on the current pose.
The transformed offset points are the new attachment point positions.

Computing which vertices are attachment points and the corresponding offsets happens before the simulation is run.
For this, we wrote a script, which can be used in the Isaac Sim GUI.
The attachment data is saved as USD properties and retrieved during the GIPC initialization.

\subsection{Tactile Simulation}
\label{subsec:tactile_sim}
\begin{figure}
    \centering
    \includegraphics[width=\textwidth]{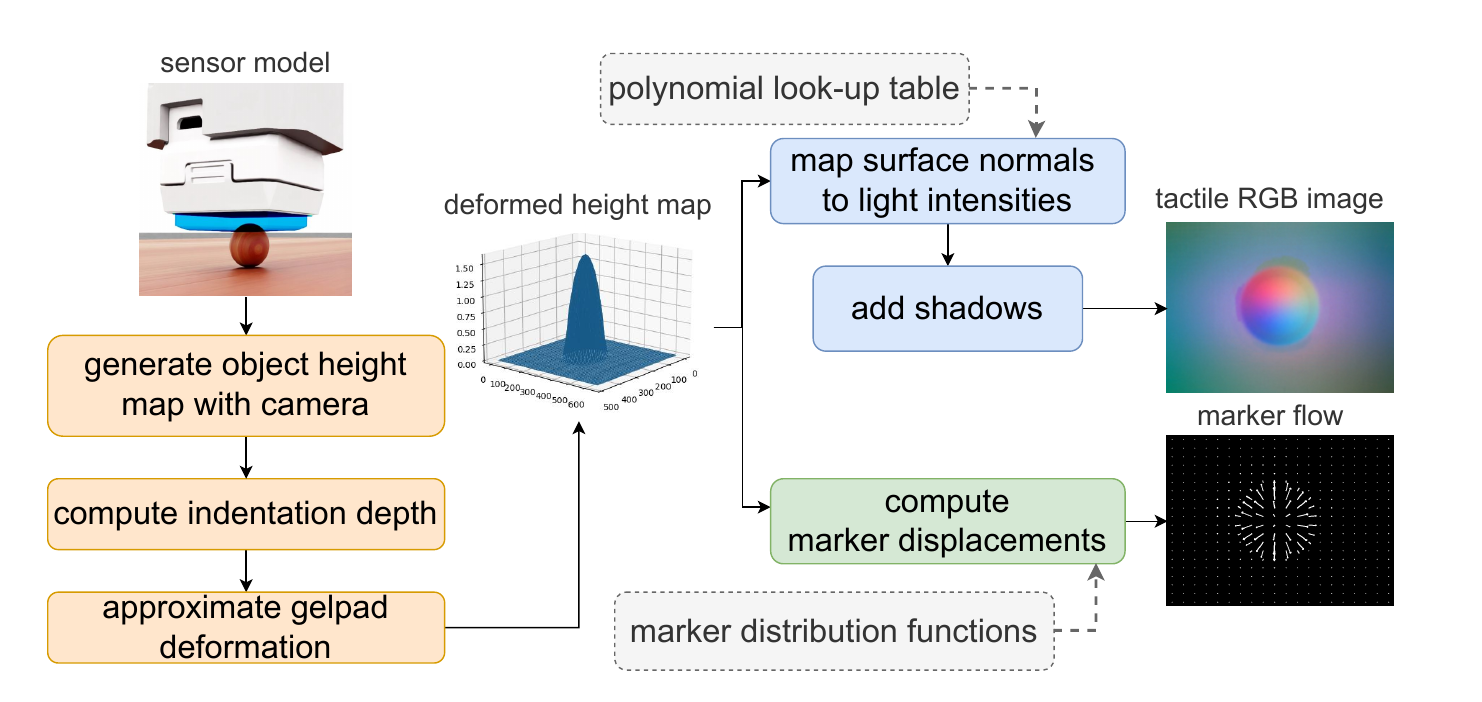}
    \caption{
    \textbf{Overview of our simulation pipeline for the sensor output.}
    We use approaches that rely on a height map which approximates the gelpad deformation.
    The height map generation is outlined in orange. 
    For generating tactile RGB images, we use Taxim~\citep{TaximExamplebasedSimulationsi2021} (blue), which uses a polynomial look-up table. For the marker flow, we use FOTS~\citep{FOTSFastOpticalzhao2024} (green), which uses functions that model the marker displacements distributions. 
    }
    \label{fig:tactile_sim_pipeline}
\end{figure}
    Our sensor model inside Isaac Sim contains a camera, which generates a height map of object surfaces inside the gelpad.
    The indentation depth is computed based on the height map and the gelpad thickness.
    Indentation depth and pyramid Gaussian kernels are used to approximate the gelpad deformation.
    For the tactile RGB image generation, the surface normals are mapped to light intensities with a polynomial look-up table.
    Then shadows are added to the RGB image to make the image more realistic.
    For the marker flow, we compute the marker displacement with exponential functions that model the marker displacements distribution under different loads. The main steps of the sensor output simulation are visualized in~\autoref{fig:tactile_sim_pipeline}.

\section{System Specification}
    We run the experiments from~\autoref{sec:demonstartions_and_eval} on an Ubuntu system with an AMD Ryzen $9$ $5950$X $16$-Core CPU, $32 \, \mathrm{GB}$ RAM and an
    NVIDIA RTX $3080$Ti GPU with $12 \, \mathrm{GB}$ VRAM.
    
\end{document}